# Définition d'une pièce test pour la caractérisation d'une machine UGV

**Kwamivi Bernardin MAWUSSI (\*)**
**Laurent TAPIE**

*Laboratoire Universitaire de Recherche en Production Automatisée*
*ENS Cachan ; 61, avenue du Président Wilson ; 94235 Cachan Cedex*
*Correspondant : anselmetti@ lurpa.ens-cachan.fr*
*(\*) Maître de conférences, IUT de St Denis, Université Paris 13*

RÉSUMÉ. *Dans plusieurs domaines comme l'aéronautique, l'outillage et l'automobile, l'usinage des pièces se fait de plus en plus sur des centres d'usinage à grande vitesse. Aujourd'hui, l'offre pour l'achat de ces centres d'usinage est très étendue. Cette situation pose le problème du choix judicieux et objectif répondant aux besoins industriels qu'il faut bien exprimer. Le choix reste difficile dans la mesure où les données techniques mises à la disposition des clients par les constructeurs de machines outils sont insuffisantes aussi bien quantitativement que qualitativement. Dans cette communication nous présentons un protocole de caractérisation des centres d'usinage en vue d'orienter le choix. Le protocole est basé d'une part sur des essais à vide complémentaires à ceux préconisés par les normes ISO 230 et ISO 10791 et d'autre part sur les essais en charge sur une pièce test. Dans la première partie, nous présentons les besoins industriels ainsi qu'une analyse des caractéristiques techniques des machines outils. La deuxième partie est consacrée à l'étude des normes, la description du protocole et la présentation des résultats.*

ABSTRACT. *In several fields like aeronautics, die and automotive, the machining of the parts is done more and more on high speed machines tools. Today, the offer for purchasing these machine tools is very wide. This situation poses the problem of the judicious and objective choice meeting industrial needs that must be necessary well expressed. The choice remains difficult insofar as the technical data provided to the customers by the manufacturers of machine tools are insufficient as well quantitatively as qualitatively. In this paper we present a protocol for the characterization of machines tools in order to direct the choice. The protocol is based on the one hand on no-load complementary tests to those recommended by the standards ISO 230 and ISO 10791 and on the other hand on the tests in load on a part test. In the first part, we present the industrial needs as well as an analysis of the technical data of machine tools. The second part is devoted to the study of the standards, the description of the protocol and the presentation of the results.*

MOTS-CLÉS *Caractérisation des machines outils, usinage grande vitesse, normes ISO d'essais des machines outils*

KEYWORDS: *Characterization of machine tools, high speed machining, ISO standards for tests of machine tools*

**1. Introduction**

Après la conception, l'élaboration de la gamme d'usinage et l'opération d'usinage elle même constituent une phase importante de l'industrialisation des produits mécaniques. Cette phase devient même cruciale dans certains domaines comme l'aéronautique, la réalisation d'outillages et l'automobile. Dans ces domaines, l'usinage à grande vitesse (UGV) est devenue rentable grâce aux avancées réalisées en matière de centres d'usinage et outils de coupe (Fallböhmer *et al.*, 2000).

Devant l'étendue de l'offre commerciale en terme de machines outil, l'industrie est actuellement confrontée à un problème de choix. En effet, les constructeurs de machines outils mettent à la disposition de leurs clients des données techniques (caractéristiques géométriques et cinématiques) insuffisantes quantitativement et/ou qualitativement pour permettre un choix objectif face aux besoins spécifiques des entreprises (Arslan *et al.*, 2004).

Pour bien orienter ce choix, il faut souvent non seulement analyser les données techniques fournies par les constructeurs mais aussi réaliser des tests de caractérisation des machines présélectionnées. Il existe des normes internationales et nationales qui décrivent les tests généraux à réaliser. Les plus connues, brièvement présentées dans cette communication sont préconisées pour la réception des machines. L'analyse des besoins industriels montrent que les résultats des tests de base réalisés à la réception des machines ne permettent pas d'y répondre. Il faut donc recourir à des tests d'évaluation des performances des machines très peu abordés dans les travaux de recherche (Wilhelm *et al.*, 1997).

Dans cette communication, nous présentons un protocole de caractérisation de centres d'UGV basé sur les essais préconisés dans les normes et l'usinage d'une pièce test initialement conçue par le Centre d'Etude Technique des Industries Mécaniques (CETIM) pour la division « pièces prismatiques de motorisation » de RENAULT (Mawussi et al, 2004).

**2. Caractéristiques des machines**

Les machines outils considérées possèdent 4 ou 5 axes. Comme pour toute machine à commande numérique, les déplacements sont assurés à partir de consignes de vitesse, elles mêmes élaborées à partir des consignes de position. Pour atteindre les vitesses, une machine passe par des phases successives d'accélération et de décélération. Précision, vitesses et accélérations constituent les premières caractéristiques qu'il faut identifier sur les machines outils d'Usinage Grande Vitesse (UGV). Pour mesurer l'importance de chaque caractéristique, il convient de la mettre en relation avec les besoins industriels.

**2.1.** *Besoins industriels*

L'analyse des contextes d'utilisation des machines outils permet de distinguer 4 secteurs ayant des besoins spécifiques (Vidal, 2004). Le premier secteur est celui de l'usinage de pièces de forme complexe comme les moules et les outillages d'emboutissage. Les machines utilisées requièrent surtout une bonne précision dynamique en suivi de trajectoires. Le deuxième secteur est celui de l'usinage dans la masse de pièces aéronautiques. Le grand débit de copeaux est une nécessité qui détermine la productivité dans ce secteur. L'usinage de pièces d'automobile caractérise le troisième secteur dans lequel le faible temps de cycle est un paramètre de productivité recherché. Le quatrième et dernier secteur est celui de la mécanique générale. La fabrication étant unitaire ou de petite série, ce secteur nécessite une grande polyvalence et une bonne ergonomie de programmation CN pour les machines utilisées.

Dans tous ces secteurs, on note également le besoin d'une bonne qualité géométrique des pièces fabriquées. L'ensemble des besoins de chaque secteur d'usinage se traduit par de fortes contraintes sur certaines caractéristiques. Ainsi, dans le cas de l'usinage de pièces automobiles, l'analyse des entités usinées permet de dégager 4 principaux critères de productivité (Tapie, 2004) : la dynamique des axes de déplacement linéaire, la répétabilité – homogénéité des axes XY, le bon synchronisme des axes Z et A pour une machine 4 axes et le temps de cycle.

**2.2.** *Exemple de fiche technique*

| Capacité en usinage | | Capacité pièce | |
|---|---|---|---|
| Course suivant X, Y, Z (mm) | 560, 400, 500 | Encombrement X, Y, Z (mm) | 900, 600, 650 |
| Déplacement rapide suivant X, Y, Z (mm/min) | 60000, 50000, 50000 | Poids maximal (kg) | 500 |
| Avance (mm/min) | 1-8500 | **Broche** | |
| Accélération (G ou ) | 0,8 | Vitesse (tr/min) | 30000 |
| Jerk (mm/s$^3$) | 100 | Puissance (kW) | 17 |
| Résolution (affichage) | 0.1 µm | Temps d'accélération (s) | 2,04 |
| Précision du déplacement | < 0.3 µm/250mm | **Changeur d'outils** | |
| Indexage axe C résolution - précision | 0,0001° - 0,003° | Nombre d'outils | 40 |
| … | | Temps de changement (s) | 2,8 |

**Table 1.** *Fiche technique d'une machine UGV*

La tableau ci-dessus (Table 1) montre la description partielle de la fiche technique d'une machine. Plusieurs caractéristiques regroupées au sein de quatre principales rubriques sont en général données sur les fiches techniques. La

« capacité pièce » fournit des informations qui doivent simplement être prises en compte par les services méthodes et planification de la production. Les trois autres rubriques présentent des caractéristiques plus ou moins directement liées aux besoins décrits plus haut et sont donc analysées dans la suite.

Dans la rubrique « capacité en usinage » sont définies les caractéristiques de la machine relatives aux différents déplacements (en travail ou rapide) et suivant les différents axes de travail. L'avance et l'accélération suivant les axes linéaires X, Y et Z interviennent fortement dans l'évaluation du temps de cycle. La valeur du jerk ainsi que la répétabilité ne sont souvent pas données. L'homogénéité et le synchronisme des axes ne peuvent être mesurés au travers des données uniaxiales.

Les caractéristiques de la broche interviennent dans la définition partielle des conditions limites de coupe. Associées aux vitesses d'avance et au choix d'autres paramètres technologiques liés à la stratégie d'usinage, ces caractéristiques influent directement sur la qualité géométrique des pièces.

Le temps de changement d'outil est la caractéristique la plus importante dans la rubrique « changeur d'outil » et il n'est souvent pas précisé sur les fiches techniques. Lorsqu'il est donné, il représente le temps non effectif car mesuré à partir de la position dite de dégagement ou de changement d'outil qui est en général très éloignée de la zone de travail courante.

*2.3. Nécessité d'une caractérisation*

La présentation des différentes rubriques disponibles sur les fiches techniques montre leur inadéquation et incomplétude par rapport aux besoins industriels. Aussi, il est difficile voire impossible de faire un choix de machine outil simplement à partir des fiches techniques fournies par les constructeurs. Il faut avant tout procéder à la caractérisation, c'est-à-dire l'évaluation des caractéristiques pertinentes des machines retenues après l'étude préliminaire des fiches techniques. Cette caractérisation passe par l'exécution de protocoles d'essais de validation des machines outils.

## 3. Essais de caractérisation des machines outils

*3.1. Normes et standards*

Plusieurs essais de validation de machines outils ont fait l'objet de normes internationales (*ISO 230* et *ISO 10791*) et nationales (*ANSI B5.54*, *BSI 4656*, *JIS B6336* et *B6201*). Notre travail s'est appuyé sur les normes internationales *ISO 230* et *ISO 10791* qui décrivent les codes et conditions d'essais pour les machines outils et centres d'usinage. Les essais de caractérisation à vide d'une part et en charge d'autre part que nous présentons dans la suite ont été élaborés à partir de ces normes. Une présentation complète des deux normes serait fastidieuse et inappropriée dans cette communication. Nous avons choisi de ne présenter que les grandes lignes relatives aux principales sources d'erreur identifiées dans la littérature. Ces sources d'erreurs sont de manière générale de six types (Rahman, 2004) : géométrie des composants et structures de la machine, température, frottement, efforts de coupe, servo commande et vibration résiduelle (dynamique).

*Tests à vide*

Les erreurs liées aux défauts géométriques des composants et structures d'une machine outil représente l'un des plus importants pourcentages des erreurs globales (Wilhelm *et al.*, 1997). La figure 1 montre les types de défauts décrits dans les normes. Chacun des défauts est soit lié à la géométrie propre d'un organe de la machine (Fig. 1a), soit lié à un mouvement relatif d'organes (Fig. 1b). Les normes *ISO 10 791-1*, *ISO 10 791-2* et *ISO 10 791-3* dérivées de la norme *ISO 230-1* donnent des indications sur les tests de conformité géométrique à réaliser sur les centres d'usinage (CU) en fraisage. Dans le cadre de nos travaux, nous n'avons pas repris ces tests. Ils sont donc considérés comme pré-requis.

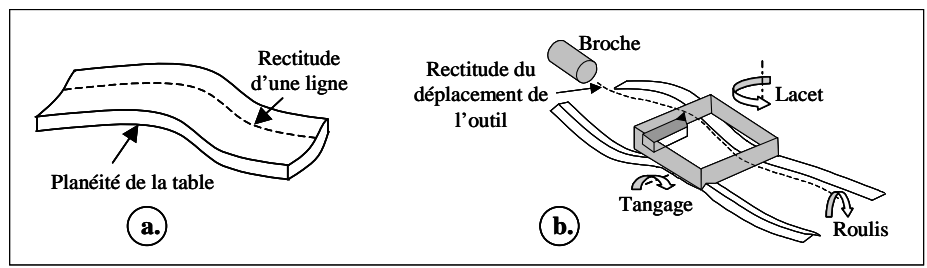

**Figure 1.** *(a)- Qualité géométrique d'une table de fraiseuse (b)- Guidage du chariot longitudinal*

Lors des déplacements avec jerk (dérivée de l'accélération) élevé les déplacements de masses mobiles génèrent des vibrations qui dégradent la qualité de l'usinage. Certaines parties des normes *ISO 230* et *ISO 10791* proposent des tests pour évaluer les défauts introduits à ce niveau. Au cours de nos travaux, le comportement dynamique engendré par les chocs et les vibrations n'a pas été évalué séparément. Le positionnement intégrant le comportement dynamique est évalué au même titre que la vitesse, l'accélération et le jerk.

Le comportement thermique est difficile à évaluer (Mutellip *et al.*, 2003). Ainsi, dans le cadre de notre travail, nous avons fait l'hypothèse que les erreurs liées aux déformations thermiques sont négligeables dans la mesure où les déplacements sont très limités dans le temps, la broche possède un système de refroidissement et la température du site expérimental est constante.

*Caractérisation en charge*

La norme *ISO 10791-7* propose un protocole de test pour caractériser le comportement d'un CU en usinage prismatique. Cette norme se réfère à la norme *ISO 230-1* qui préconise une série d'essais de coupe dans des conditions de finition de pièces standards. Les entités ainsi que les spécifications minimales définies sur les pièces doivent permettre de déterminer la précision de l'usinage de la machine considérée. Deux types de pièces ayant deux tailles différentes sont proposées.

La première pièce consiste : en un positionnement, à l'alésage de cinq trous et en une série de passes de finition sur différents profils. L'objectif de cette première pièce est de contrôler les performances du C.U. dans différentes conditions cinématiques comme par exemple l'avance suivant un axe unique, l'interpolation linéaire et circulaire impliquant deux axes. La seconde pièce sert à contrôler la planéité d'un plan usiné par surfaçage de finition grâce à deux passes se chevauchant d'une valeur de 20% du diamètre de la fraise.

Des indications pour les paramètres de coupe sont fournis, ainsi que quelques indications peu précises pour la géométrie des outils. Les valeurs conseillées sont proches de l'usinage conventionnel ce qui pose le problème du passage à l'UGV. Quelques recommandations sont données sur la phase d'ébauche notamment l'importance d'une surépaisseur d'usinage constante pour la passe de finition.

### 3.2. Essais réalisés

Le protocole que nous proposons se décompose en 3 phases de déplacements libres des axes numériques, une simulation d'usinage à vide, une simulation d'usinage en charge. Ce protocole a été expérimenté sur un centre d'usinage MIKRON UCP 710 équipé d'un directeur de commande numérique Siemens 840D. Les relevés de positions ont été effectués à l'aide du servo trace intégré au DCN. Pour les relevés temporels nous avons utilisés un sous-programme relevant les données de l'horloge interne du DCN et stockant les relevés dans les registres du DCN. L'ensemble de ces données a ensuite été traité à l'aide d'une Macro développée sous Excel. Les relevés dimensionnels sur la pièce usinée ont été effectués à l'aide d'une Machine à Mesurer Tridimensionnelle (MMT).

*Déplacements à vide*

Ces essais ont pour objectifs de caractériser la dynamique, la précision, la répétabilité en positionnement et l'homogénéité des axes machines. Les trajets d'outils sont définis en fonction des courses de la machine (Fig. 2). Les conditions et modes de sollicitation de la machine sont également précisés dans la figure 2.

Pour évaluer la dynamique des axes de la machine, nous avons retenu comme indicateurs les valeurs maximales de la vitesse d'avance, l'accélération et le jerk. Les valeurs relevées sont présentées dans la première partie de la table 2. Un déplacement linéaire uni axial ou interpolé proche de la valeur de la course permet à la machine d'atteindre la vitesse rapide de consigne. Sur les tracés obtenus (non présentés), nous avons noté une instabilité au niveau des accélérations et jerks. L'instabilité est beaucoup plus importante sur l'axe Y qui a une plus forte dynamique (20%) que l'axe X. Les résultats relatifs au trajet A2 A5 mettent en évidence la bonne homogénéité des deux axes X et Y. Nous avons rencontré des problèmes de coupure de puissance sur l'axe Z. Les résultats de l'essai réalisé à 80% de la consigne sur cet axe concordent avec ceux obtenus sur les axes X et Y.

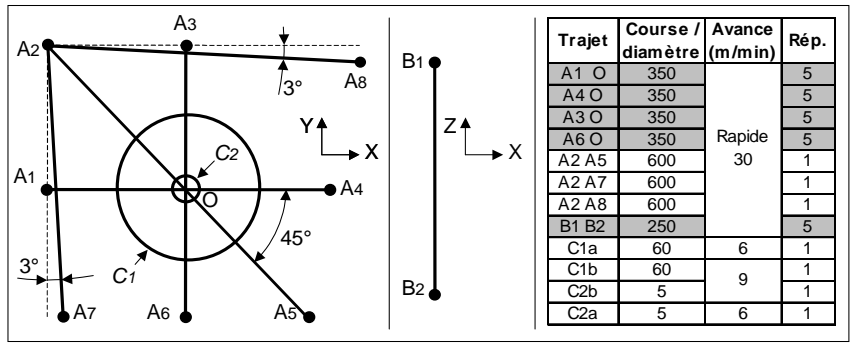

**Figure 2.** *Trajectoires du test à vide*

| | Trajet | Axe | Vitesse (m/min) | Accél. (m/s²) | Jerk (m/s³) |
|---|---|---|---|---|---|
| *Dynamique globale* | A1 O | X | 30,29 | 2,73 | 119,85 |
| | A4 O | X | 30,33 | 2,69 | 119,89 |
| | A3 O | Y | 30,4 | 3,35 | 143,43 |
| | A6 O | Y | 30,42 | 3,31 | 143,45 |
| | A2 A5 | X | 30,3 | 2,73 | 121,04 |
| | | Y | 30,32 | 2,77 | 119,32 |
| | A2 A7 | X | 1,59 | 0,18 | 7,16 |
| | | Y | 30,42 | 3,28 | 142,51 |
| | A2 A8 | X | 30,3 | 2,72 | 120,48 |
| | | Y | 1,59 | 0,15 | 6,65 |
| | B1 B2 | Z | 24,19 [1] | 2,33 | 108,55 |
| | C1a | X | 6,01 | 0,68 | 8,68 |
| | | Y | 6,01 | 0,37 | 5,39 |
| | C1b | X | 9,01 | 0,83 | 13,77 |
| | | Y | 9,01 | 0,77 | 7,69 |
| | C2b | X | 3,21 | 0,59 | 12,68 |
| | | Y | 3,21 | 0,59 | 9,42 |
| | C2a | X | 3,21 | 0,6 | 12,19 |
| | | Y | 3,21 | 0,59 | 8,97 |

| Axe | Ecart position (µm) | Répétabilité (µm) | Réversibilité (µm) |
|---|---|---|---|
| X $^{(+)}$ | 20,12 | 2,29 | 12,4 |
| X $^{(-)}$ | 7,72 | 1,37 | |
| X $^{(+/-)}$ | 13,92 | 19,71 | |
| Y $^{(+)}$ | 39,78 | 0,72 | 10,58 |
| Y $^{(-)}$ | 29,2 | 1,3 | |
| Y $^{(+/-)}$ | 34,49 | 14,6 | |
| Z $^{(+)}$ | -- [2] | -- [2] | -- [2] |
| Z $^{(-)}$ | 3,66 | 0,96 | |
| Z $^{(+/-)}$ | -- [2] | -- [2] | |
| ***Précision des axes linéaires*** | | | |

| ***Précision des interpolations circulaires*** | | | |
|---|---|---|---|
| **Cercle** | **Circularité (µm)** | **Ecart radial maxi (µm)** | **Ecart radial mini (µm)** |
| C1a | 29,03 | 22,09 | -5,87 |
| C1b | 57,16 | 48,46 | -1,6 |
| C2a | 42,97 | 35,52 | -3,3 |
| C2b | 43,21 | 36,23 | -3,27 |

**Table 2.** *Résultats des essais à vide.* [1] *Test réalisé à 80% de la consigne en raison de la coupure de la puissance.* [2] *Test impossible à réaliser pour la même raison.*

Pour les interpolations circulaires, le trajet C2 confirme le phénomène du fort ralentissement et saturation déjà observé au niveau des interpolations linéaires. Les valeurs de l'accélération et du jerk restent également très limitées.

La précision des déplacements linéaires est évaluée au travers de trois indicateurs : l'écart de position (consigne – réelle) du point d'arrivée, la répétabilité et la réversibilité (écart entre deux directions d'approche). Les écarts de position et les répétabilités sont évalués suivant deux approches unidirectionnelles (+) ou (-) et une approche bidirectionnelle (+/-). Les résultats présentés dans la deuxième partie de la table 2 montrent que l'axe X a un meilleur positionnement et une moins bonne répétabilité que l'axe Y. La réversibilité quasi identique sur les deux axes X et Y reste élevée. Cette mauvaise réversibilité des axes a une forte influence sur le positionnement et la répétabilité bidirectionnels. Les problèmes rencontrés sur l'axe Z n'ont pas permis de réaliser les tests de précision. Pour un industriel, ces problèmes peuvent remettre en cause la bonne réception de la machine, même si les tests normalisés ont été concluant. Suivant les recommandations des normes, seul le positionnement de l'axe Z peut être validé.

Dans le cas des interpolations circulaires, nous avons retenu les deux indicateurs de précision suivant proposés dans les normes : l'écart de circularité et l'écart radial (entre une position réelle et le cercle nominal de consigne). Les résultats des essais sur les deux cercles C1 et C2 avec deux vitesses différentes (Fig. 2) sont présentés dans la troisième partie de la table 2. Malgré les valeurs élevées, les écarts de circularité et les écarts mini radiaux correspondent aux recommandations des normes.

La spécificité de l'UGV non prévue dans les normes se retrouve à deux niveaux. Le premier concerne les résultats obtenus pour le positionnement des axes X et Y, la répétabilité et la réversibilité de tous les axes. Le deuxième niveau est relatif aux écarts maxi radiaux mesurés pour les interpolations circulaires.

*Essais en charge*

Dans notre protocole, les tests en charge ont été réalisé sur la pièce dont les entités géométriques sont présentées dans la figure 3. Il s'agit de la pièce initialement conçue par le CETIM pour RENAULT. La pièce test se compose d'entités d'usinage typiquement rencontrées sur les culasses, blocs moteurs et carters de boîte de vitesse. Dans le protocole, la géométrie est figée non seulement pour la pièce finie mais aussi pour le brut (non présenté) qui est fourni. Afin de constituer un référentiel des machines testées, RENAULT fournit également les outils de coupe, un porte pièce et le programme d'usinage. Dans le cadre de nos travaux, nous avons non seulement utilisé ces différents éléments du protocole mais aussi contribué à leur amélioration et la mise au point de nouvelles versions. L'ensemble de la nouvelle version du protocole sera présenté dans une prochaine communication.

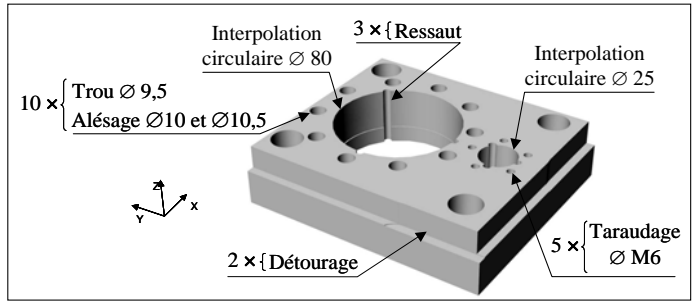

**Figure 3.** *Géométrie de la pièce test*

La première partie de la caractérisation en charge porte sur le test d'usinage à vide (sans la pièce brute). Cette étape intermédiaire permet à la fois de vérifier le comportement dynamique de la machine et créer une référence des indicateurs temporels indépendants des efforts de coupe.

Les résultats du comportement dynamique de la machine sont présentés dans la table 3. Les relevés relatifs aux détourages sont conformes à ceux obtenus en déplacement uni axial libre. Pour les alésages, la prépondérance dynamique de l'axe Y par rapport à l'axe X est également confirmée.

|  |  | Vitesse (m/min) | | Accélération (m/s²) | | Jerk (m/s3) | |
|---|---|---|---|---|---|---|---|
|  |  | Axe X | Axe Y | Axe X | Axe Y | Axe X | Axe Y |
| **Détourages** | *Consigne* | 2,87 | 2,87 | 0,8 | 0,81 | 64,92 | 63,77 |
|  | *Réel* | 2,88 | 2,89 | 0,46 | 0,47 | 19,53 | 16,25 |
| **Alésages Ø25** | *Consigne* | 2,01 | 2,01 | 0,48 | 0,52 | 47,59 | 97,42 |
|  | *Réel* | 2,03 | 2,02 | 0,49 | 0,48 | 17,95 | 21,56 |
| **Alésages Ø80** | *Consigne* | 9,55 | 9,55 | 0,85 | 0,85 | 35,12 | 60,93 |
|  | *Réel* | 9,57 | 9,56 | 0,86 | 0,9 | 16,71 | 19,57 |

**Table 3.** *Comportement dynamique en usinage à vide*

Dans deuxième partie de la caractérisation en charge, la pièce test a été usinée. Les indicateurs temporels et la qualité géométrique de la pièce usinée ont été évalués dans cette partie. La dynamique n'a plus été retenue comme indicateur à ce niveau en raison de l'absence d'informations complémentaires apportées par les tests de l'usinage à vide.

Les valeurs relevées pour les indicateurs temporels dans les deux parties de la caractérisation en charge sont données dans la table 4. La différence entre les temps de finition des détourages confirme celle déjà constatée au niveau des dynamiques des axes X et Y. On note un gain de temps de 25% sur l'axe Y, ce qui est très significatif dans le temps de cycle. Au niveau des alésages (réalisés en interpolation circulaire) les temps sont quasi identiques pour le diamètre 25 à cause du phénomène de fort ralentissement et saturation. Pour le diamètre 80, les temps sont différents. Entre 530 et 750 m/min, la machine franchit un palier de vitesse de rotation dont la dynamique de la broche a une forte influence sur celle des axes linéaires. Cette interaction entre vitesse de rotation de la broche et avance des axes linéaires sera étudiée au cours de travaux futurs. Les temps des perçages-alésages et taraudages constituent des références mais n'apportent pas d'informations complémentaires. Enfin, la comparaison entre les temps d'usinage à vide et d'usinage de la pièce permettent de conclure sur l'absence d'influence de la coupe sur les résultats des essais en charge.

| Entité / phase | Temps (s) | Entité / phase | Temps (s) | Entité / phase | Temps (s) |
|---|---|---|---|---|---|
| Ebauche des détourages | 7,93 [1]<br>**7,95 [2]** | Alésage 2 ⌀25<br>580 m/min | 0,82<br>**0,82** | Taraudage 1<br>20 m/min | 4,93<br>**5** |
| Finition détourage YZ | 2,58<br>**2,545** | Alésage 3 ⌀25<br>670 m/min | 0,81<br>**0,81** | Taraudage 2<br>40 m/min | 5,03<br>**5,07** |
| Finition détourage XZ | 3,46<br>**3,47** | Alésages ⌀25<br>Cycle global | 8,21<br>**8,21** | Taraudage 3<br>60 m/min | 4,85<br>**5,15** |
| Alésage 1 ⌀80<br>530 m/min | 2,19<br>**2,2** | Perçages ⌀9,5<br>Cycle global | 32,17<br>**32,16** | Taraudage 4<br>80 m/min | 5,24<br>**5,46** |
| Alésage 2 ⌀80<br>750 m/min | 1,73<br>**1,77** | Alésages ⌀10<br>Cycle global | 34,33<br>**34,39** | Taraudage 5<br>100 m/min | 5,34<br>**5,31** |
| Alésage 3 ⌀80<br>940 m/min | 1,66<br>**1,63** | Alésages ⌀10,5<br>Cycle global | 24,62<br>**24,65** | Changement d'outil | 12,22<br>**12,19** |
| Alésages ⌀80<br>Cycle global | 17,39<br>**17,21** | Perçages ⌀5<br>Cycle global | 10,98<br>**11** | Cycle global | 295,6<br>**295,08** |
| Alésage 1 ⌀25<br>470 m/min | 0,81<br>**0,81** | | | | |

**Table 4.** *Indicateurs temporels [1] pour l'usinage à vide [2] pour l'usinage de la pièce*

La qualité géométrique de la pièce test usinée est évaluée au niveau des entités qui la composent. Les indicateurs qualitatifs et quantitatifs ainsi retenus sont présentés ci-dessous.

– Alésages ⌀80 et ⌀25 réalisés en interpolation circulaire : nous avons mesuré les défauts de circularité. Ce type de spécification est souvent rencontré dans le secteur automobile, en particulier au droit des portés de roulement des boîtes de vitesse.
– Détourages : nous avons d'une part mesuré les défauts de planéité de chaque surface finie ainsi leur orientation relative (perpendicularité) et d'autre part évalué les écarts de distance (nominale – réelle) entre les profils ébauche et finition. L'indicateur relatif aux écarts de distance permet de caractériser les déformations de l'ensemble {coulant, porte outil, outil}.

– Alésages ⌀10 et ⌀10,5 réalisés à partir du réseau des dix avant trous ⌀9,5 : nous avons mesuré les défauts de localisation qui caractérisent le positionnement de la machine.

Les résultats des mesures et évaluations sont donnés dans la table 5. Pour les alésages ⌀25, les défauts de circularité sont identiques. L'écart observé sur le premier alésage est dû à un effet de bord, car il a été supprimé après correction de la position en Z de l'entité. L'augmentation progressive des vitesses de coupe et d'avance lors de l'usinage des alésages ⌀80 a engendré des effets dynamiques dont l'influence se retrouve dans la mesure des défauts de circularité. L'effet de palier déjà constaté est confirmé à ce niveau.

| | Entité | Circularité (µm) | Entité / phase | Planéité (µm) | Coulant (µm) | Localisation des alésages ⌀10 et ⌀10,5 (µm) |
|---|---|---|---|---|---|---|
| Interpolation circulaire | Alésage ⌀80 530 m/min | 26 | Détourage YZ ébauche | 37 | 16 mini 41 maxi | Minimum : 48 Maximum : 70 Moyenne : 64,3 Dispersion (6σ) : 41,2 |
| | Alésage ⌀80 750 m/min | 56 | Détourage YZ finition | 7 | | |
| | Alésage ⌀80 940 m/min | 71 | Détourage XZ ébauche | 4 | -20 mini 33 maxi | |
| | Alésage ⌀25 470 m/min | 35 | Détourage XZ finition | 8 | | |
| | Alésage ⌀25 580 m/min | 20 | ***Détourages + Coulant*** | | | ***Positionnement*** |
| | Alésage ⌀25 670 m/min | 20 | **Perpendicularité des détourages finition** (µm) | 13 | | |

**Table 5.** *Qualité géométrique des entités de la pièce usinée*

La mesure du défaut de planéité sur les surfaces détourées confirme simplement l'importance des perturbations introduites par la dynamique de l'axe Y. Les écarts calculés mettent en évidence une légère déformation de l'ensemble {coulant, porte outil, outil} confirmée par la mesure du défaut de perpendicularité. Les défauts de localisation mesurés sur le réseau de trous sont très importants surtout lorsqu'on les compare à la précision des axes linéaires caractérisée au cours des tests de déplacement libre (voir table 2). La moyenne des défaut de localisation égale à 64,3 µm doit être rapportée à l'écart maxi de position qui vaut 39,78 µm. Quant à la dispersion de 41,2 µm, elle est comparée avec réserve à la valeur maximale de répétabilité égale à 19,71 µm. La différence entre ces deux types de résultats en partie due à la déformation des alésoirs qui ont un faible diamètre, une grande longueur (de 80 à 120 mm) et sont creux (lubrification au centre).

Face aux besoins du secteur automobile et aux critères de productivité choisis, les essais de caractérisation de la machine MIKRON mettent en évidence finalement des perturbations liées à la dynamique de l'axe Y, une précision des axes linéaires assez bonne, un comportement en interpolation circulaire dégradé par la dynamique des axes, une bonne homogénéité des axes et une qualité géométrique moyenne.

**4. Conclusion**

Un protocole de caractérisation des machines outils UGV à vide et en charge a été proposé dans cette communication. Au cours des essais à vide, des tests complémentaires à ceux définis dans les normes de base ont été élaborés. L'usinage d'une pièce test, précédé d'un usinage à vide, est au centre des essais en charge.

Le protocole a été réalisé sur le centre d'UGV MIKRON de notre laboratoire. Les résultats obtenus permettent de caractériser la machine par rapport aux besoins industriels exprimés sous forme de critères de productivités. Par ailleurs, ces résultats sont intégrés à un référentiel qui favorise le choix d'une machine au travers de leur comparaison. Les différentes informations ainsi recueillies peuvent également être utilisées par les services méthodes dans la génération des gammes d'usinage.

**5. Bibliographie**